\documentclass[pdflatex,sn-mathphys-num]{sn-jnl}

\usepackage{graphicx}%
\usepackage{multirow}%
\usepackage{amsmath,amssymb,amsfonts}%
\usepackage{amsthm}%
\usepackage{mathrsfs}%
\usepackage[title]{appendix}%
\usepackage{xcolor}%
\usepackage{textcomp}%
\usepackage{manyfoot}%
\usepackage{booktabs}%
\usepackage{algorithm}%
\usepackage{algorithmicx}%
\usepackage{algpseudocode}%
\usepackage{listings}%
\usepackage{placeins} 
\usepackage{paralist}
\usepackage{tikz}
\usepackage{subcaption}
\usetikzlibrary{arrows.meta,positioning,shapes.misc,fit,backgrounds}

\theoremstyle{thmstyleone}%
\theoremstyle{thmstyletwo}%

\theoremstyle{thmstylethree}%

\begin{document}

\title[Article Title]{Enhancing Cross-Patient Generalization in AI-Based Parkinson’s Disease Detection}


\author*[1]{\fnm{Mhd Adnan} \sur{Albani}}\email{mohammad.adnan@safee.com}

\author[2]{\fnm{Riad} \sur{Sonbol}}\email{riad.sonbol@hiast.edu.sy}

\affil*[1]{\orgname{Safee technologies Company}, \orgaddress{\city{Dubai}, \country{United Arab Emirates}}}

\affil[2]{\orgdiv{Department of Informatics,}, \orgname{Higher Institute for Applied Sciences and Technology (HIAST)}, \orgaddress{\city{Damascus},\country{Syria}}}


\abstract{Parkinson's disease (PD) is a neurodegenerative disease affecting about 1\% of people over the age of 60, causing motor impairments that impede hand coordination activities such as writing and drawing. Many approaches have tried to support early detection of Parkinson's disease based on hand-drawn images; however, we identified two major limitations in the related works: (1) the lack of sufficient datasets, (2) the robustness when dealing with unseen patient data. In this paper, we propose a new approach to detect Parkinson's disease that consists of two stages: The first stage classifies based on their drawing type(circle, meander, spiral), and the second stage extracts the required features from the images and detects Parkinson's disease. We overcame the previous two limitations by applying a chunking strategy where we divide each image into 2×2 chunks. Each chunk is processed separately when extracting features and recognizing Parkinson's disease indicators. To make the final classification, an ensemble method is used to merge the decisions made from each chunk. Our evaluation shows that our proposed approach outperforms the top performing state-of-the-art approaches, in particular on unseen patients. On the NewHandPD dataset our approach, it achieved $97.08\%$ accuracy for seen patients and $94.91\%$ for unseen patients, our proposed approach maintained a gap of only 2.17 percentage points, compared to the 4.76-point drop observed in prior work.}



\maketitle

\section{Introduction}\label{sec1}
Neurological disorders have become a primary cause of disability around the world. Parkinson's Disease (PD) is the fastest-growing neurological disorder in the world \cite{bib5}. From 1990 to 2015, the number of PD patients doubled to reach over 6 million \cite{bib20}, and studies say that by 2040, more than 12 million people could be diagnosed with PD \cite{bib1}.

\begin{figure}[t]
    \centering
    \begin{subfigure}{0.13\linewidth}
        \includegraphics[width=\linewidth]{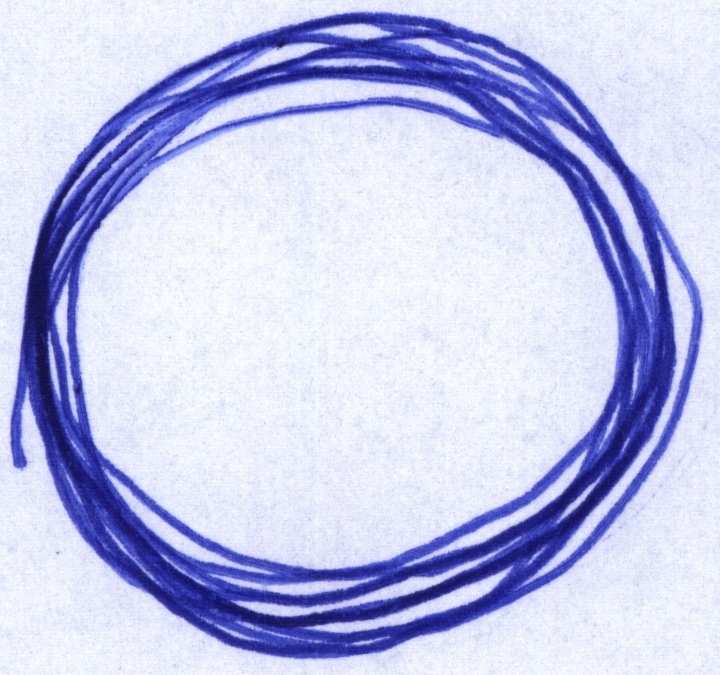}
        \caption*{a1}
    \end{subfigure}
    \begin{subfigure}{0.13\linewidth}
        \includegraphics[width=\linewidth]{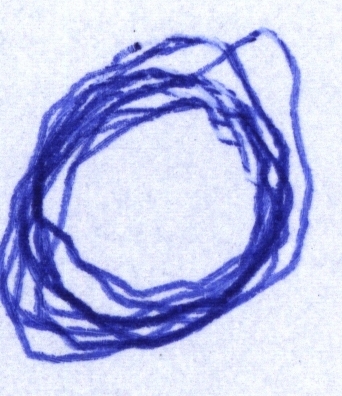}
        \caption*{a2}
    \end{subfigure}
    \hspace{0.4cm}
    \begin{subfigure}{0.13\linewidth}
        \includegraphics[width=\linewidth]{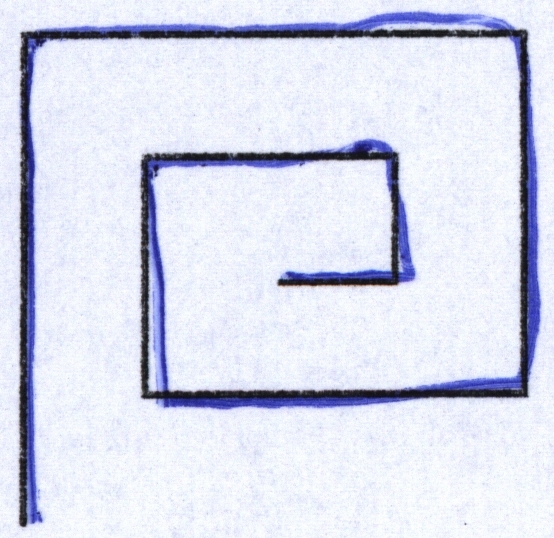}
        \caption*{b1}
    \end{subfigure}
    \begin{subfigure}{0.13\linewidth}
        \includegraphics[width=\linewidth]{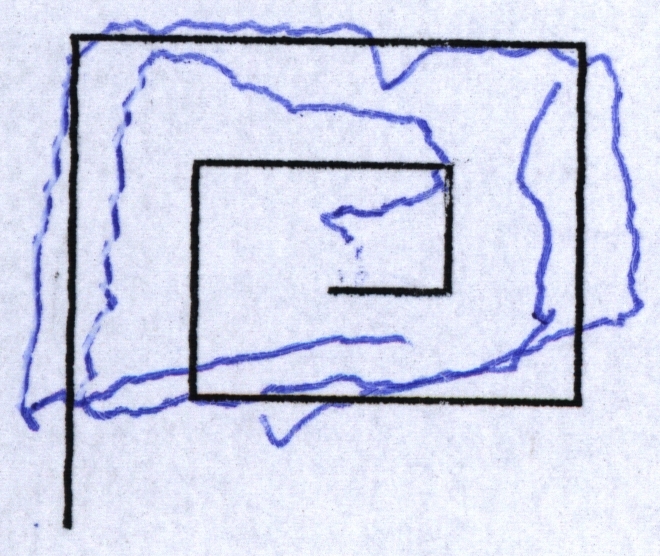}
        \caption*{b2}
    \end{subfigure}
    \hspace{0.4cm}
    \begin{subfigure}{0.13\linewidth}
        \includegraphics[width=\linewidth]{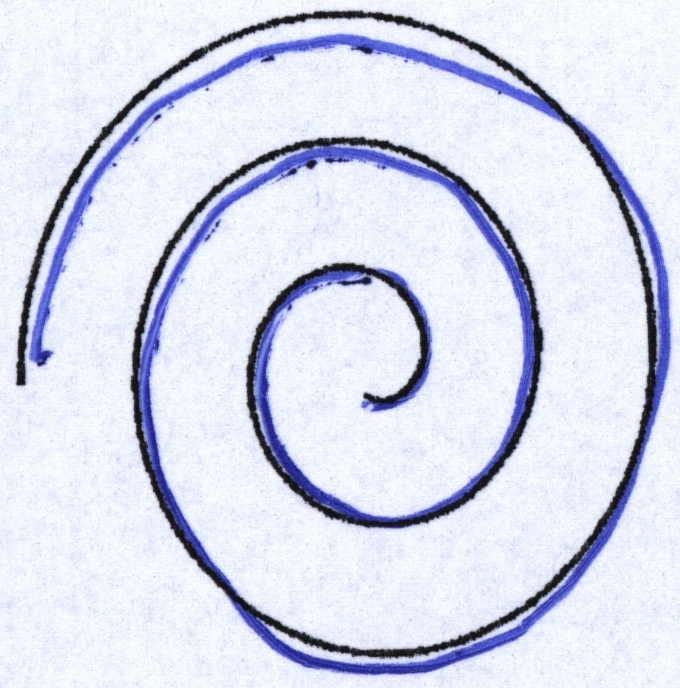}
        \caption*{c1}
    \end{subfigure}
    \begin{subfigure}{0.13\linewidth}
        \includegraphics[width=\linewidth]{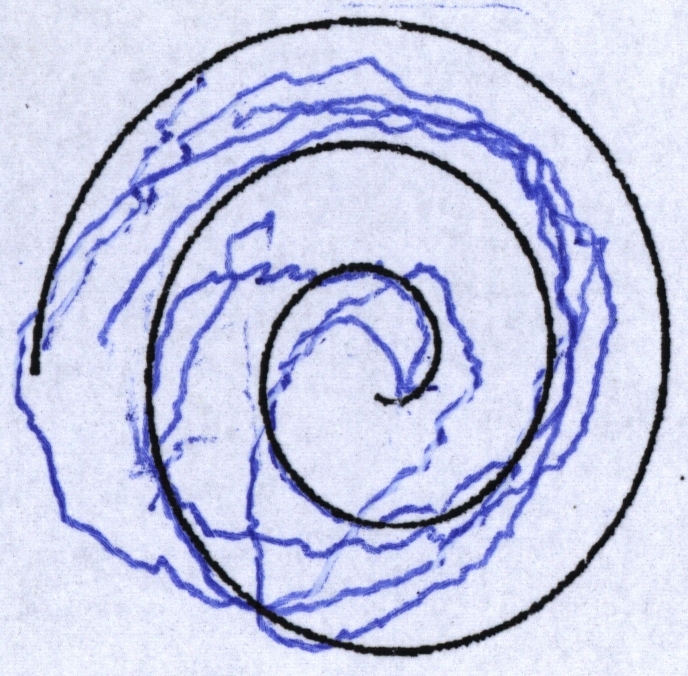}
        \caption*{c2}
    \end{subfigure}

    \caption{Examples of hand-drawing samples.}
    \label{fig:drawings_horizontal}
\end{figure}

Parkinson's Disease effects roughly $1\%$ of the population over the age of 60 \cite{bib3}. It causes motor impairments, such as tremors, rigidity, and sluggishness \cite{bib2}, which in turn impact fine motor skills, including drawing and writing \cite{bib4}. Since drawing requires both cognitive function and fine motor control, it is suitable for early PD diagnosis \cite{bib7}. This has prompted the development of drawing-based diagnostics, such as the hand-drawing assessment, which includes tasks like drawing circles, spirals, and meanders, to evaluate fine motor skills (as shown in Figure~\ref{fig:drawings_horizontal}) \cite{bib21}.

Figure \ref{fig:drawings_horizontal} Shows the different types of drawing tasks: circle, meander, and spiral. These tasks are used to evaluate motor performance and assist in diagnosing individuals with Parkinson’s disease from healthy individuals.

Traditional assessments have inspired the development of machine learning methods that evaluate drawing patterns to identify PD. These approaches offer a noninvasive and affordable evaluation system, providing a cheap alternative to conventional clinical evaluations \cite{bib10}. Moreover, they can uncover features that are relevant to PD detection that might go unnoticed in clinical diagnosis \cite{bib8}.

Several studies have applied machine learning approaches to detect PD from hand drawings. Some of these works used Classical machine learning approaches, such as a decision tree classifier (DT), K-nearest-neighbor(KNN), Support-Vector-Machine(SVM) \cite{bib11,bib16}, while others used more advanced machine learning approaches, such as a Convolutional neural network (CNN) \cite{bib9,bib12,bib14}, ensemble methods  \cite{bib13}, and multi-modal approaches\cite{bib15}.

Although these results are promising, there remains a gap between the reported accuracies in the papers and their ability to generalize to unseen data. In many of the existing works, the same individual has multiple drawing samples. When these samples are randomly divided into training and testing sets, data from the same individual can appear in both. This introduces data leakage, causing the model to learn individual-specific features rather than patterns of PD. This, in turn, will lead to optimistic performance estimates. 

Given the limitations in the evaluation criteria for drawing-based PD detection, this study aims to address these gaps, guided by the following research questions:

\textbf{RQ1: How can we evaluate the generalization of PD detection?}

This question focuses on the identification of suitable metrics and experimental setups to evaluate model generalization.

\textbf{RQ2: How can we design a more robust approach towards unseen data?}

This question focuses on developing techniques to enhance the model's ability to generalize when dealing with unseen data.

To address these questions, we propose a solution that incorporates an individual-wise evaluation strategy to assess the models' ability to generalize to unseen subjects. Also, we introduce a chunking-based approach that divides each hand-drawn image into smaller segments (chunks or tiles) to extract localized features, thereby mitigating the challenge posed by the limited dataset size. A feature map will represent each chunk, which is then classified using classical machine learning techniques. The final prediction for each image is determined through a majority voting scheme across all chunk-level predictions.

The remaining sections of this paper is structured as follows: 
Section~\ref{sec:related} reviews related works on AI-based PD detection. 
Section~\ref{sec:method} describes the proposed methodology.
Section~\ref{sec:experiment} presents the experimental study, setup, and evaluation protocol. 
Section~\ref{sec:results} Presents the results and compares them with existing approaches.
Section~\ref{sec:AnswersToRQ} discusses the findings and provides answers to the research questions.
Section~\ref{Sec:Threats} outlines the threats to validity.
Finally, Section~\ref{sec:conclusion} concludes with key contributions, future directions, and threats to validity.

\section{Relates Works}\label{sec:related}
Recent studies have employed a wide range of Artificial intelligence and machine learning approaches for detecting PD using hand-drawn data. These approaches use a broad range of models, techniques, and evaluation strategies. Below, we review prior work categorized by methodology, discussing their main contributions, outcomes, and limitations.

\subsection{Classical Machine learning Approach}

Parziale et al. \cite{bib11} conducted one of the early comparative studies on machine learning methods for Parkinson's disease (PD) detection using offline hand-drawing samples. Using the NewHandPD dataset, which contains circle, spiral, and meander drawings (as shown in Figure~\ref{fig:drawings_horizontal}) from PD patients and healthy controls, the study evaluated three classical classifiers: Decision Tree (DT), Support Vector Machine (SVM), and Random Forest (RF). The results showed that the Decision Tree achieved higher accuracy compared to the SVM while maintaining greater interpretability, as it could represent decisions through explicit if–then rules familiar to Medical Practice. Moreover, the Random Forest achieved the best overall accuracy, but its decision process was less interpretable. This study found that a decision tree can adequately detect PD from drawings while providing clear understandability. However, due to the handcrafted features and the small-sized dataset, the model's ability to capture an adequate representation was limited, whereas deep learning methods can automatically capture such representations.

Rios-Urrego et al \cite{bib23}. examined PD patient samples by combining kinematic, geometrical, and non-linear features. In feature extraction, a Wacom tablet was used; these features include velocity, acceleration, pressure, and entropy, from spiral and sentence tasks. This study includes 149 individuals, consisting of 55 PD patients and the remaining 94 healthy subjects. Classical machine learning techniques, such as SVM, KNN, and RF classifiers, were in this tested study, the techniques achieved an accuracy of up to $93.1\%$. Validation of the models was performed on an independent dataset, achieving an accuracy of $83.3\%$, which shows promising generalization performance.

\subsection{CNN based Approach}
With the rise of deep learning, Convolutional neural networks (CNNs) have become a common approach for PD detection. Khatamino et al. \cite{bib16} proposed an approach that employs a custom CNN that was trained on spiral drawings as well as the Dynamic Spiral Test (DST). Their CNN architecture is composed of of two convolutional layers and two max-pooling layers, followed by two fully connected layers with ReLU activations and a final Softmax output layer. This approach achieved $88\%$ accuracy. These results highlight the ability of CNNs to learn directly from raw hand-drawn features. The approach was evaluated using K-fold cross-validation and Leave-One-Out cross-validation (LOOCV). As expected, LOOCV produced a lower and more variable accuracy ($72\%$ - $88\%$), indicating a high sensitivity to validation choice . Although the results indicate the practicality of this method, the study is limited by its small dataset (72 subjects with an imbalance of the control group). Furthermore, while both K-fold and LOOCV were used for evaluation, only LOOCV is explicitly individual-wise, as for the K-fold, it's unclear whether it's Image-wise or subject-wise split. This ambiguity limits confidence in the reported performance.

Farhah \cite{bib19} investigated the use of transfer learning models, VGG19 \cite{bib28}, InceptionV3 \cite{bib29}, ResNet50v2 \cite{bib30}, and DenseNet169 \cite{bib25} pretrained on a dataset consisting of 102 spirals (equally split between PD and healthy subjects). InceptionV3 achieved the best performance among these models reaching $89\%$ accuracy. While the results are promising, the relatively small dataset, as well as being restricted to a single type of drawing task, may limit the extent to which the findings can be generalized to a broader clinical settings.

Chakraborty et al. \cite{bib17} extended this line of work by proposing a dual-stage CNN classifier, which was trained on a dataset that includes spiral and wave drawings from both PD patients and healthy subjects. The accuracy achieved by their system was $93\%$ outperforming Classical ML approaches. While these results show the ability of CNN in capturing features useful for identifying PD, the evaluation methodology relied on image-wise cross-validation, where multiple samples from the same subject could appear in train and test sets. This choice risks data leakage and is likely to result in an overly optimistic performance estimate, raising concerns about the true generalizability.

\subsection{Ensemble Methods}
Rai et al. \cite{bib13} proposed a weightedd ensemble method that combined three CNNs, DenseNet121 \cite{bib25}, MobileNetV2 \cite{bib26}, and NASNetMobile \cite{bib27}. Each model's contribution to the final output was determined using a performance-based weighting scheme. Particularly, each CNN was evaluated individually and then assigned a weight proportional to its accuracy. This strategy enabled the ensemble method to achieve up to $95\%$ accuracy on spiral and $90\%$ on wave drawings.

\subsection{Hierarchical Approach}
Kansizoglou et al. \cite{bib18} Proposed a hierarchical deep learning approach to identify PD from hand-drawing tasks (circles, spirals, meanders). The drawing is analyzed in two stages. The first stage focuses on drawing type classification (circles, spirals, meanders), and the second stage employs a dedicated model for PD detection specific to each drawing type. The authors' proposed architecture achieved an accuracy of $93.6\%$ for circle, $96.7\%$ for meander, and $97.9\%$ for spiral with an overall accuracy of $96.79\%$, outperforming conventional CNN baselines. Although these results were promising, the evaluation was based on image-wise cross-validation, this highlights limitations in the model's performance of generalization and real-world performance estimates.

\begin{table}[htpb]
\caption{Comparison of Classical and Deep Learning Approaches for PD Detection from Hand Drawings}\label{tab1}%
\begin{tabular}{@{}lllll@{}}
\toprule
\textbf{Study} & \textbf{Representation} & \textbf{Classification} & \textbf{Evaluation Type} & \textbf{Accuracy} \\
\midrule
Rios-Urrego et al. \cite{bib23} & Feature-based & Classical ML & Subject-wise & 83.3\% \\

\multirow{2}{*}{Khatamino et al. \cite{bib16}} & \multirow{2}{*}{CNN-based} & \multirow{2}{*}{Fully Connected} & Image-wise & 88\% \\ 
 &  &  & Subject-wise & 72\% \\

\multirow{1}{*}{Chakraborty et al. \cite{bib17}} & CNN-based & Fully Connected & Image-wise CV & 93.3\% \\

\multirow{1}{*}{Rai et al. \cite{bib13}} & CNN-based & Fully Connected & Image-wise CV & 95\% (spiral), 90\% (wave) \\

\multirow{1}{*}{Kansizoglou et al. \cite{bib18}} & CNN-based & Fully Connected & Image-wise CV & 96.97\% \\

Farhah \cite{bib19} & CNN-based & Fully Connected & Image-wise CV & 89\% \\
\botrule
\end{tabular}
\end{table}

\noindent{
\paragraph{Disscussion:} Table~\ref{tab1} compares classical and deep learning approaches for PD detection from hand drawings. Classical feature-based models, such as Rios-Urrego et al.~\cite{bib23}, achieved moderate accuracy ($83.3\%$) under subject-wise evaluation, demonstrating robust generalization to unseen data. In contrast, CNN-based methods reported higher accuracies (up to $97\%$), as seen in Chakraborty et al.~\cite{bib17}, Rai et al.~\cite{bib13}, and Kansizoglou et al.~\cite{bib18}. However, most of these evaluations were done using image-wise, which may inflate performance due to potential data overlap between subjects. Khatamino et al.~\cite{bib16} showed a clear decrease from $88\%$ (image-wise) to $72\%$ (subject-wise), highlighting this issue. Overall, while CNNs outperform classical models in raw accuracy, subject-wise evaluations reveal that their generalization to unseen individuals remains limited.}

\section{Methodology}\label{sec:method}
To overcome the mentioned challenges, our methodology is based on the main ideas. First, we increase the number and diversity of the training data by applying data augmentation to solve data limitation issues. Secondly, we introduce the concept of image chunking, where each drawing is divided into a 2x2 grid. This process mimics the natural way humans analyze drawings. Humans tend to focus on specific areas first to spot irregularities, and then gradually build an overall understanding of the drawing. More importantly, chunking acts as an ensemble-like mechanism that helps reduce the variance of ML models, which is a core factor affecting generalization, by aggregating predictions from multiple localized regions of the same drawing. Finally, using an individual-wise data split ensures that the samples from the same individual do not appear in both the training and testing sets, preventing data leakage and resulting in a more realistic estimate of performance. 

In the following section, we will discuss the proposed methodology for Parkinson's disease (PD) detection from hand-drawing tasks.
In particular it describes \textbf{a)} Preprocessing and Data Augmentation, \textbf{b)} The proposed two-stage DL-based approach. \textbf{c)} Training, \textbf{d)} Evaluation protocol.

\subsection{Data Preprocessing and Augmentation}\label{augmentationt}

All of the experiments apply the same preprocessing pipelines. To ensure reproducibility across the experiments, the preprocessing follows a fixed order of operations consisting of four steps: resizing, tiling, normalization, and deterministic augmentation.   

\begin{itemize}
    \item \textbf{Resizing:}  
    Each image is resized to $448 \times 448$ pixels.

    \item \textbf{Chunking:}  
    A fixed $2 \times 2$ grid partition is applied to the $448 \times 448$ image, producing four non-overlapping tiles of size $224 \times 224$ pixels each.

    \item \textbf{Normalization:}  
    Each tile is normalized per channel so that the brightest pixel in each channel reaches a value of one.

    \item \textbf{Deterministic Data Augmentation:}  
    Before chunking, augmentations are applied on the resized image $\hat{I}$, with the drawing type determining repeat counts.  
    Let $r_d$ indicate the number of repeats for drawing type $d \in \{\text{circle}, \text{meander}, \text{spiral}\}$, where:
    \[
    r_{\text{circle}} = 4, \quad r_{\text{meander}} = 2, \quad r_{\text{spiral}} = 2.
    \]

    \begin{itemize}
        \item For \textbf{circle} drawings, each repeat corresponds to a fixed rotation angle:
        \[
        \theta_k = \frac{360^\circ}{r_{\text{circle}}} \cdot k, \quad k=0,1,2,3,
        \]
        using bilinear interpolation with white padding.

        \item For \textbf{meander} and \textbf{spiral} drawings, zero-mean Gaussian noise with a standard deviation of $\sigma=0.003$ is added:
        \[
        \tilde{I}(x,y) = \text{clip}\left( \hat{I}(x,y) + \mathcal{N}(0,\sigma^2) \right), \quad \tilde{I} \in [0,1].
        \]
    \end{itemize}

    Noise is seeded deterministically using a SHA-256 hash of the draw type, file path, and repeat index, ensuring reproducibility.
\end{itemize}

\subsection{Model Architecture}
We propose a three-stage architecture for PD detection.
The architecture is broken down into sequential stages: Drawing-type classification, Feature extraction, and PD classification. 
This design promotes modularity, allowing each component to be optimized independently.

\paragraph{Stage 1: Drawing-Type Classification.}  
In this stage, a ResNet-based encoder \(h_{\phi}\) \cite{bib31} is used to identify the drawing type (circle, meander, or spiral) from the preprocessed image \(X\). The model follows the standard residual-learning design, consisting of a \(7 \times 7\) convolution and max-pooling layer, followed by four residual stages with \(3 \times 3\) convolutions. Each stage increases the number of feature channels (64, 128, 256, 512) and uses skip connections to preserve information across layers. A global average pooling layer and a fully connected Softmax layer produce the final prediction:
\[
d = h_{\phi}(X), \quad d \in \{\text{circle}, \text{meander}, \text{spiral}\}.
\]
This ResNet-based architecture enables effective feature extraction and stable training for drawing-type classification.

\paragraph{Stage 2: Feature Extraction.}\label{sec:feature_extraction}
Given the same input image \( X \), a feature extractor \( f_{\theta_d} \) generates a feature vector:
\[
z = f_{\theta_d}(X) \in \mathbb{R}^m.
\]
We evaluated three types of feature extraction models:
\begin{itemize}
    \item \textbf{ResNet-based encoder \cite{bib31}}: CNN backbone with residual connections.
    \item \textbf{Pyramid Vision Transformer (PVT) \cite{bib32}}: Pure Transformer backbone with a CNN-like pyramid, producing multi-scale features across four stages.
    \item \textbf{Hybrid model (ResNet + PVT)}: Concatenates both embeddings to form a fused feature vector consisting of 1024 dimensions, which is fed to the downstream classifier.
\end{itemize}

All feature extractors were initialized with ImageNet-pretrained weights and fine-tuned on the training set to adapt to the characteristics of hand-drawn PD data.

\paragraph{Stage 3: PD Classification.}
The extracted feature vector \( z \) is passed to a classifier \( c_d \) to produce the final PD prediction:
\[
\hat{y} = c_d(z), \quad \hat{y} \in \{\text{PD}, \text{Healthy}\}.
\]
We evaluated multiple machine learning classifiers:
\begin{itemize}
    \item \textbf{$k$-Nearest Neighbors (KNN)} classifies samples based on spatial similarity. Performs well on low-dimensional, well-separated embeddings such as those extracted from hand-drawn segments, where each tile encodes local geometric cues for detecting PD patients and healthy controls.
    
    \item \textbf{Decision Tree (DT)} offers interpretable, rule-based decisions through hierarchical feature splits. This transparency is valuable for understanding which drawing traits contribute to PD detection.
    
    \item \textbf{Random Forest (RF)} combines multiple decision trees to reduce overfitting and improve generalization, making it well-suited for small datasets like the one used.
    
    \item \textbf{Neural Network (NN)} captures non-linear relationships between features, enabling the model to learn complex patterns.
\end{itemize}

Together, these cover a spectrum from interpretable rule-based learners to flexible non-linear classifiers, enabling a comprehensive evaluation of how well the extracted features generalize to unseen subjects.

\begin{figure}[b]
  \centering
  \includegraphics[width=\linewidth]{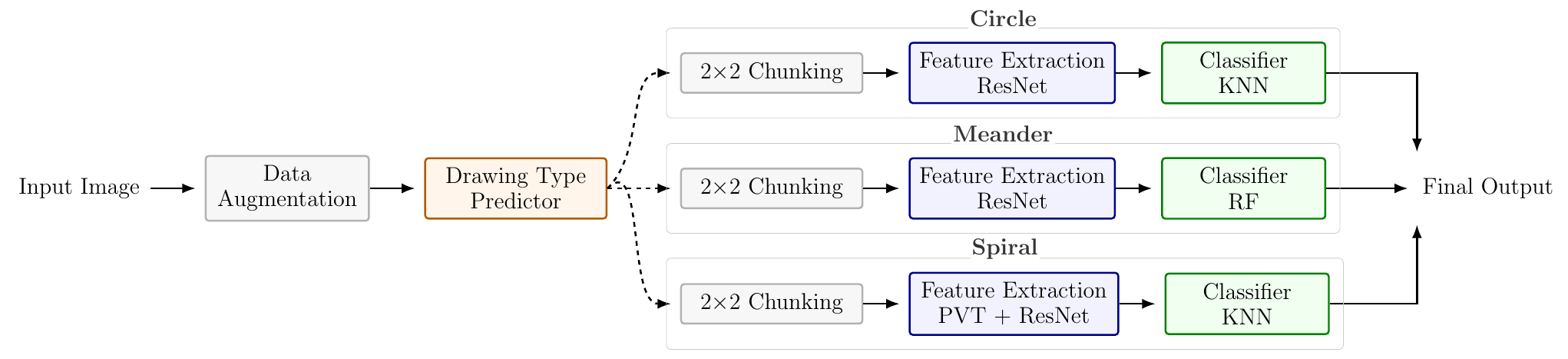}
  \caption{Proposed two-stage architecture for drawing-aware PD detection.}
  \label{fig:arch-diagram}
\end{figure}

\vspace{2 em}

\section{Experiments and results}\label{sec:experiment}

\subsection{Experiment setup}

In this section we present the experimentation setup for the proposed three-stage architecture, covering the dataset, implementation variations, and Evaluation metrics.

\subsubsection{Dataset}

The dataset used in the experiments is the \textit{NewHandPD}~\cite{bib22} dataset, which consists of 279 hand-drawn shapes retrieved from 66 individuals. Thirty-five individuals are healthy, while 31 individuals are diagnosed with Parkinson's disease (PD).

For each individual, the following drawings are available:
\begin{itemize}
    \item \textbf{Circle:} 1 drawing per individual,
    \item \textbf{Meander:} 4 drawings per individual,
    \item \textbf{Spiral:} 4 drawings per individual.
\end{itemize}

This results in \(9\) drawings per person and a total of \(66 \times 9 = 594\) images across the dataset.
Accordingly, the healthy group contributes \(35 \times 9 = 315\) images. Figure \ref{fig:drawings_horizontal} shows some examples of the drawings from the used dataset.

\subsubsection{Evaluation Methodology}

To evaluate the proposed solution, multiple evaluation strategies were explored:

\begin{itemize}
   \item \textbf{5-fold image-wise CV (Img-CV5): } This strategy randomly splits the dataset into five parts (folds) at the \textit{image level} irrespective of the individual identity. 
    In each iteration, four parts(folds) are used for training and one for testing, so that every image is tested exactly once across the five runs. 
    Since drawings from the same individual may appear in both training and testing sets, this setup is used to illustrate the potential generalization gap compared to individual-wise evaluation.

    \item \textbf{5-fold Individual-wise CV (Ind-CV5): } This strategy is similar to the previous one, with a key difference being that the split is done at the individual level, ensuring that all the samples from a single individual appear only in either the training or the testing set. This strategy evaluates the ability of the model to generalize to unseen subjects.  preventing any subject overlap between the training and testing set.

    \item \textbf{leave-one-individual-out cross-validation (LOIO): } To examine the robustness under limited data availability, the leave-one-out strategy was employed. In each iteration, one individual is held out for testing, while the remaining data are used for training.
\end{itemize}

\subsubsection{Evaluation Metrics}

The dataset includes three drawing types: \textit{Circle}, \textit{Meander}, and \textit{Spiral}. 
Each drawing type was evaluated separately using \textbf{True Positives (TP)}, \textbf{False Positives (FP)}, \textbf{True Negatives (TN)}, \textbf{False Negatives (FN)}, 
\textbf{Accuracy}, \textbf{Precision}, \textbf{Recall}, and \textbf{F1-Score}. 
To obtain an overall measure of performance across all drawings, we computed the \textbf{weighted accuracy}, which represents the total accuracy over the entire dataset while accounting for the number of samples per drawing type. 
It is defined as:
\begin{equation}
\label{eq:weighted_accuracy}
A_{\text{weighted}} = 
\frac{\displaystyle\sum_{i=1}^{D} A_i \, N_i}{\displaystyle\sum_{i=1}^{D} N_i}, 
\quad D = 3 \; (\text{Circle, Meander, Spiral}),
\end{equation}
where \( A_i \) and \( N_i \) denote the accuracy and number of samples for drawing type \( i \), respectively. 
This ensures that \( A_{\text{weighted}} \) corresponds to the overall accuracy across all drawings.

\subsubsection{Tested Configurations}\label{sec:tested_config}
To examine the impact of architectural and design choices, several configurations were evaluated. These included:
\vspace{-1 em}
\begin{itemize}
    \item \textbf{Backbone architectures:} Several options were tested such (shown in \ref{sec:feature_extraction}) as, ResNet, PVT, and a hybrid ResNet+PVT representations by concatenating the features from both models.
    \item \textbf{Chunking strategy:} We tested various chunking options such as no-chunking, 2×2 and 3×3 Chunking schemes.
    
    \item \textbf{Classification Method:} We compared several ML approaches, specifically we tested, $k$-Nearest Neighbors (KNN), Random Forest (RF), Decision Tree (DT), and Neural Network(NN).
\end{itemize}

\subsubsection{Hardware Setup}
All experiments were conducted in Python~3.10.18 using PyTorch~2.4.0 with CUDA~12.1. 
The system featured an NVIDIA RTX~4070~Ti~SUPER (16~GB VRAM), an Intel Core~i9-13900K CPU, and 64~GB RAM. 
Model architectures and pretrained weights were sourced from torchvision~(v0.19.0) and timm~(v1.0.19). 
Training was performed deterministically with fixed random seeds to ensure reproducibility.

\subsection{Results}\label{sec:results}
This section presents the key experimental results for the proposed approach for (PD) detection. Several combinations of backbones and classifier were tested under different chunking and data augmentation settings, as outlined in \textbf{Section~\ref{sec:tested_config}}.

\subsubsection{Drawing Type classifier Results}
The first stage of the proposed solution is classifying images into their drawing type (i.e., Circle, Meander, or Spiral). This task is considered an easy task since each drawing type has its own distinctive features. As expected, this step achieved a perfect accuracy of \textbf{100\%}, confirming that the drawings were easily distinguishable and that the following PD detection results were not affected. Since this task was relatively straightforward, it is not analyzed further in this section.

\subsubsection{PD Detector Results}
Here, we will present the results of the PD detection stage. \textbf{First} we will present the results of the best performing setups based on the three evaluation strategy (Img-CV5, Ind-CV5, LOIO) . \textbf{Second} we will compare the result of our approach to recent studies based on Img-CV5 since this is the strategy that is used among different research papers. \textbf{Finally}. A detailed comparison will be made with the top performing state-of-the-art solution using the three evaluation strategies. Later in \textbf{Ablation Study (Section~\ref{sec:ablation})}, a detailed overview of all experimental configurations and their comparative analysis is provided.

\paragraph{A. Best-performing setups.}

The top results for each drawing type (i.e., Circle, Meander, and Spiral) are summarized in Table \ref{tab:resnet_best_results}. Results show that ResNet was the best backbone for Circle and Meander, while for spiral, we achieved the best results using a concatenated representation of PVT and ResNet. Across all experiments, the \textbf{2×2 chunking} strategy consistently resulted in the best performance. 
Using the full image captures the drawing as one feature map, while chunking splits it into smaller parts, which are then processed separately.
This lets each part use the model’s full representational power to learn local details and stroke variations that might be missed when analyzing the whole image at once. 
This idea mimics the way humans analyze drawing for PD patterns.
Additionally, applying \textbf{data augmentation}  (as defined in section \ref{augmentationt}) further boosted performance by increasing the number of training samples in a relatively small dataset.

\begin{table}[!h]
\centering
\caption{Best results and configuration details for CV5 Image-wise evaluation.}
\label{tab:resnet_best_results_imagewise}

\scriptsize
\setlength{\tabcolsep}{3pt}
\renewcommand{\arraystretch}{1.0}

\begin{tabular}{@{}llcccccccccc@{}}
\toprule
\textbf{BB} & \textbf{Draw/Cls} & \textbf{Acc} & \textbf{Prec} & \textbf{Rec} & \textbf{F1} & \textbf{TP} & \textbf{FP} & \textbf{TN} & \textbf{FN} & \textbf{Aug} & \textbf{Chnk} \\ 
\midrule

\multirow{2}{*}{\textbf{ResNet}} 
 & Circ–KNN & \textbf{0.939} & 0.889 & 1.000 & 0.941 & 32 & 4 & 30 & 0 & Yes & 2×2 \\
 & Meand–RF & \textbf{0.973} & 0.969 & 0.977 & 0.973 & 125 & 4 & 132 & 3 & Yes & 2×2 \\
 \\

\textbf{PVT+ResNet} & Spir–KNN & \textbf{0.977} & 0.953 & 1.000 & 0.977 & 128 & 6 & 130 & 0 & Yes & 2×2 \\
\midrule

\multicolumn{11}{r}{\textbf{Weighted Avg:}} & \textbf{97.08} \\

\bottomrule
\end{tabular}
\end{table}

\begin{table}[!h]
\centering
\caption{Best results and configuration details for CV5 Individual-wise evaluation.}
\label{tab:resnet_best_results}

\scriptsize
\setlength{\tabcolsep}{3pt}
\renewcommand{\arraystretch}{1.0}

\begin{tabular}{@{}llcccccccccc@{}}
\toprule
\textbf{BB} & \textbf{Draw/Cls} & \textbf{Acc} & \textbf{Prec} & \textbf{Rec} & \textbf{F1} & \textbf{TP} & \textbf{FP} & \textbf{TN} & \textbf{FN} & \textbf{Aug} & \textbf{Chnk} \\ 
\midrule

\multirow{2}{*}{\textbf{ResNet}} 
 & Circ–KNN & \textbf{0.939} & 0.889 & 1.000 & 0.941 & 32 & 4 & 30 & 0 & Yes & 2×2 \\
 & Meand–RF & \textbf{0.958} & 0.953 & 0.961 & 0.957 & 123 & 6 & 130 & 5 & Yes & 2×2 \\
 \\

\textbf{PVT+ResNet} & Spir–KNN & \textbf{0.943} & 0.925 & 0.961 & 0.943 & 123 & 10 & 126 & 5 & Yes & 2×2 \\
\midrule

\multicolumn{11}{r}{\textbf{Weighted Avg:}} & \textbf{94.91} \\

\bottomrule
\end{tabular}
\end{table}

\vspace{-3em}

\begin{table}[!h]
\centering
\caption{Best results and configuration details for LOIO evaluation.}
\label{tab:resnet_best_results_imagewise_loso}

\scriptsize
\setlength{\tabcolsep}{3pt}
\renewcommand{\arraystretch}{1.0}

\begin{tabular}{@{}llcccccccccc@{}}
\toprule
\textbf{BB} & \textbf{Draw/Cls} & \textbf{Acc} & \textbf{Prec} & \textbf{Rec} & \textbf{F1} & \textbf{TP} & \textbf{FP} & \textbf{TN} & \textbf{FN} & \textbf{Aug} & \textbf{Chnk} \\ 
\midrule

\multirow{2}{*}{\textbf{ResNet}} 
 & Circ–KNN & \textbf{0.864} & 0.848 & 0.875 & 0.862 & 28 & 5 & 29 & 4 & Yes & 2×2 \\
 & Meand–RF & \textbf{0.932} & 0.952 & 0.922 & 0.937 & 118 & 6 & 130 & 10 & Yes & 2×2 \\
 \\

\textbf{PVT+ResNet} & Spir–KNN & \textbf{0.932} & 0.930 & 0.930 & 0.930 & 119 & 9 & 127 & 9 & Yes & 2×2 \\
\midrule

\multicolumn{11}{r}{\textbf{Weighted Avg:}} & \textbf{92.40} \\

\bottomrule
\end{tabular}
\end{table}

\noindent
As shown in Table~\ref{tab:resnet_best_results_imagewise}, the \textbf{CV5 Image-wise} evaluation resulted in the highest performance, with a weighted average accuracy of \textbf{97.08\%}. 
\noindent
In comparison, \textbf{CV5 Individual-wise} setup (Table~\ref{tab:resnet_best_results}) achieved a slightly lower  \textbf{94.91\%}, showing a minimal drop of only \textbf{2.17\%} and confirming strong generalization. 
in the most challenging \textbf{LOIO} evaluation strategy (Table~\ref{tab:resnet_best_results_imagewise_loso}), 
accuracy decreased to \textbf{92.40\%}.
\medskip

The \textbf{Circle} task was consistently the most difficult, this is 
likely due to its smaller sample size and limited motion diversity, which can reduce discriminative patterns compared to Meander and Spiral. 
Overall, the results demonstrate that the proposed framework maintains high performance across all evaluation strategies, validating its reliability for early PD detection.

\paragraph{B. Comparison with prior work.}

To contextualize the performance of the proposed framework, Table~\ref{tab:comparison_simplified} compares it against recent state-of-the-art approaches for PD detection using  the HandDrawnPD dataset. This comparison is done based on Image-wise Split (Img-wise) since all works use this strategy in evaluation.

\begin{table}[h!]
\centering
\caption{Comparison of Existing Studies and the Proposed Approach.}
\label{tab:comparison_simplified}
\begin{tabular}{@{}lll@{}}
\toprule
\textbf{Method} & \textbf{Drawing Type(s)} & \textbf{Accuracy (\%)} \\
\midrule
CNN (DST) \cite{bib16} & Spiral & 79.64 \\
Transfer Learning \cite{bib19} & Multiple & 89.0 \\
SVM, KNN, RF \cite{bib23} & Multiple & 83.3 \\
Dual-stage CNN \cite{bib17} & Spiral & 93.0 \\
Hierarchical CNN \cite{bib18} & Circle, Meander, Spiral & 96.97 \\
\textbf{Proposed (Ours)} & \textbf{Circle, Meander, Spiral} & \textbf{97.08} \\
\bottomrule
\end{tabular}
\footnotetext{Accuracies are reported as stated in the original works; overall averages were not always provided.}
\end{table}

\noindent
The proposed approach achieved \textbf{97.08}, which is slightly better than the state-of-the-art results \textbf{96.97}, Although the improvement appears insignificant, it will become noteworthy in the next section, where the individual-wise evaluation highlights a much larger performance gap.

\paragraph{C. Detailed state-of-the-art comparison.}
\noindent
In this section, we provide a comprehensive comparison between the proposed hierarchical framework and existing state-of-the-art approach (Hierarchical CNN \cite{bib18}). 
The comparison employs all three evaluation strategies—\textbf{CV5 Image-wise}, \textbf{CV5 Individual-wise}, and \textbf{Leave-One-Individual-Out (LOIO)}.
This analysis highlights the robustness of our method under increasingly strict validation conditions, 
demonstrating its ability to generalize to unseen individuals.

\begin{table}[!h]
\centering
\caption{Comparison of our proposed approach and state-of-the-art \cite{bib18} results under different evaluation strategies.}
\label{tab:sota_comparison}

\scriptsize
\setlength{\tabcolsep}{4pt}
\renewcommand{\arraystretch}{1.0}

\begin{tabular}{@{}l|cc|cc|cc@{}}
\toprule
\multirow{2}{*}{\textbf{Drawing}} & 
\multicolumn{2}{c|}{\textbf{Image-wise}} & 
\multicolumn{2}{c|}{\textbf{Individual-wise}} & 
\multicolumn{2}{c}{\textbf{LOIO}} \\ 
\cmidrule(lr){2-7}
 & \textbf{Ours} & \textbf{SOTA} & \textbf{Ours} & \textbf{SOTA} & \textbf{Ours} & \textbf{SOTA} \\
\midrule
Circle  & \textbf{93.90} & 93.68 & \textbf{93.9} & 86.36 & \textbf{86.4} & 86.36 \\
Meander & \textbf{97.30} & 96.72 &\textbf{95.8} & 92.42 & \textbf{93.2} & 87.88 \\
Spiral  & 97.70 & \textbf{97.98} & \textbf{94.3} & 93.56 & \textbf{93.2} & 86.36 \\
\midrule
\textbf{Weighted Avg.} & \textbf{97.08} & 96.97 & \textbf{94.91} & 92.21 & \textbf{92.40} & 87.03 \\
\bottomrule
\end{tabular}
\end{table}

\noindent
As shown in Table~\ref{tab:sota_comparison}, our method achieves comparable results across all evaluation strategies, with the most notable advantage emerging under stricter testing conditions. 
While the state-of-the-art baseline~\cite{bib18} exhibits a notable drop in accuracy when moving from Image-wise to LOIO evaluation. 
Our approach maintains a consistently high performance. 
This smaller generalization gap demonstrates the robustness of the proposed hierarchical framework, 
which better captures individual-independent handwriting characteristics and mitigates overfitting to individual writing styles.

\section{Ablation Study}\label{sec:ablation}

To measure the impact of different design choices, we performed a series of ablation experiments. 
The goal was to quantify the effect of \textbf{chunking} and \textbf{data augmentation} on the performance of the proposed approach.
All of the experiments were performed under the same evaluation protocol (CV5 Ind-wise) and using the best-performing backbones and classifiers from Section~\ref{sec:results}. 

\vspace{0.5em}
\noindent
\textbf{1) Without Chunking.}
Table~\ref{tab:No_Chunking} shows the results when the chunking step was removed, meaning the full drawing was processed as a single image. 
The performance decreased across all drawing types, with a weighted average accuracy of only \textbf{89.71\%}. 
This highlights the importance of localized feature learning — chunking enables the model to capture fine-grained spatial variations and subtle motor irregularities that are difficult for the model to capture in global representations.

\begin{table}[!h]
\centering
\caption{Without Chunking}
\label{tab:No_Chunking}

\scriptsize
\setlength{\tabcolsep}{3pt}
\renewcommand{\arraystretch}{1.0}
\begin{tabular}{@{}llcccccccccc@{}}
\toprule
\textbf{BB} & \textbf{Draw/Cls} & \textbf{Acc} & \textbf{Prec} & \textbf{Rec} & \textbf{F1} & \textbf{TP} & \textbf{FP} & \textbf{TN} & \textbf{FN} & \textbf{Aug} & \textbf{Chnk} \\ 
\midrule
\multirow{2}{*}{\textbf{ResNet}} 
 & Circ–KNN & \textbf{0.879} & 0.875 & 0.875 & 0.875 & 28 & 4 & 30 & 4 & Yes & No \\
 & Meand–RF & \textbf{0.909} & 0.882 & 0.938 & 0.909 & 120 & 16 & 120 & 8 & Yes & No \\
\textbf{PVT+ResNet} & Spir–KNN & \textbf{0.890} & 0.878 & 0.898 & 0.888 & 115 & 16 & 120 & 13 & Yes & No \\
\midrule
\multicolumn{11}{r}{\textbf{Weighted Avg:}} & \textbf{89.71} \\
\bottomrule
\end{tabular}
\end{table}

\vspace{0.5em}
\noindent
\textbf{2) Without Data Augmentation.}
Table~\ref{tab:No_Augmentationt}  shows the results when no augmentation techniques were applied. 
Even though accuracy remained relatively high, the overall weighted average decreased noticeably compared to the fully configured solution. 
Data augmentation improved the stability and generalization of the model. This is due to artificially enlarging the training set, reducing overfitting, and helping the model handle inter-subject variability common in PD handwriting data.

\begin{table}[!h]
\centering
\caption{Without Data Augmentation}
\label{tab:No_Augmentationt}

\scriptsize
\setlength{\tabcolsep}{3pt}
\renewcommand{\arraystretch}{1.0}
\begin{tabular}{@{}llcccccccccc@{}}
\toprule
\textbf{BB} & \textbf{Draw/Cls} & \textbf{Acc} & \textbf{Prec} & \textbf{Rec} & \textbf{F1} & \textbf{TP} & \textbf{FP} & \textbf{TN} & \textbf{FN} & \textbf{Aug} & \textbf{Chnk} \\ 
\midrule
\multirow{2}{*}{\textbf{ResNet}} 
 & Circ–KNN & \textbf{0.879} & 0.853 & 0.906 & 0.879 & 29 & 5 & 29 & 3 & No & 2×2 \\
 & Meand–RF & \textbf{0.932} & 0.910 & 0.953 & 0.931 & 122 & 12 & 124 & 6 & No & 2×2 \\
\textbf{PVT+ResNet} & Spir–KNN & \textbf{0.905} & 0.899 & 0.906 & 0.903 & 116 & 13 & 126 & 12 & No & 2×2 \\
\midrule
\multicolumn{11}{r}{\textbf{Weighted Avg:}} & \textbf{90.7} \\
\bottomrule
\end{tabular}
\end{table}

\vspace{0.5em}
\noindent
\textbf{3) Without Chunking and Data Augmentation.}
Finally, Table~\ref{tab:No_Both} shows the results when both chunking and augmentation were removed. 
This configuration represents the simplest configuration, showing the combined degradation when neither chunking nor data augmentation is used. 
As expected, the overall accuracy dropped significantly to \textbf{88.3\%}, confirming that both \textbf{chunking} and \textbf{data Augmentation} are essential for robust PD detection and subject-level generalization.

\begin{table}[!h]
\centering
\caption{Without Chunking and Data Augmentation}
\label{tab:No_Both}

\scriptsize
\setlength{\tabcolsep}{3pt}
\renewcommand{\arraystretch}{1.0}
\begin{tabular}{@{}llcccccccccc@{}}
\toprule
\textbf{BB} & \textbf{Draw/Cls} & \textbf{Acc} & \textbf{Prec} & \textbf{Rec} & \textbf{F1} & \textbf{TP} & \textbf{FP} & \textbf{TN} & \textbf{FN} & \textbf{Aug} & \textbf{Chnk} \\ 
\midrule
\multirow{2}{*}{\textbf{ResNet}} 
 & Circ–KNN & \textbf{0.833} & 0.839 & 0.812 & 0.825 & 26 & 5 & 29 & 6 & No & No \\
 & Meand–RF & \textbf{0.864} & 0.854 & 0.867 & 0.860 & 111 & 19 & 117 & 17 & No & No \\
\textbf{PVT+ResNet} & Spir–KNN & \textbf{0.917} & 0.884 & 0.953 & 0.917 & 122 & 16 & 120 & 6 & No & No \\
\midrule
\multicolumn{11}{r}{\textbf{Weighted Avg:}} & \textbf{88.3} \\
\bottomrule
\end{tabular}
\end{table}

\vspace{-0.5em}
\noindent
\textbf{Discussion: }
The ablation results clearly show that both \textbf{chunking} and \textbf{data augmentation} significantly contribute to performance. 
Chunking allows localized pattern recognition that mirrors how human experts analyze fine motor control, 
while data augmentation increase data size and diversity helping with generalization and reducing overfitting. 
This combination leads to the highest overall weighted accuracy (\textbf{94.91\%}), confirming the effectiveness of the proposed design.

\section{Answers to the Research Questions}\label{sec:AnswersToRQ}

\paragraph{RQ1: How can we evaluate the generalization of PD detection?}
Our experiments show that \textbf{subject-wise} cross-validation (CV5) is a good indicator for generalization estimates. Results under image-wise CV substantially overstate performance compared to subject-wise CV due to subject overlap (data leakage). Throughout the study, we therefore report and select models using a subject-wise CV. In addition, reporting per-drawing results (Circle/Meander/Spiral) alongside a \textbf{weighted average} provides a balanced summary of overall system behavior.

\paragraph{RQ2: How can we design a more robust approach toward unseen data?}
We improve generalization in our approach through three key aspects: \textbf{(1)} a \textbf{chunk-based representation}, which divides drawings into smaller chunks or tiles, in turn allowing the model to learn localized motion irregularities, and \textbf{(2)} \textbf{deterministic data augmentation}, which increases data diversity through controlled rotations and noise. Combined, these aspects enhance the robustness of the approach, as well as make the system more consistent across evaluation strategies and less prone to overfitting.

\section{Threats to validity}\label{Sec:Threats}

While the proposed approach demonstrates strong performance, several aspects may limit the generalizability and robustness of the approaches. How ever steps were taken to reduce these threats to validity. The following points outline key threats to validity that should be considered when interpreting the findings:

\begin{itemize}
\item[(i)] \textit{Augmentation Bias}: The Deterministic rotations or noise-based augmentations can introduce artificial patterns which the model may learns to exploit, rather than truly learning Parkinson’s disease (PD)-relevant cues. However, the data augmentation in this study was carefully done to produce images that closely resemble the natural drawing style.

\item[(ii)] \textit{Drawing-Type Coverage}: Like other works, the proposed approach currently supports three drawing tasks (circle, meander, spiral). Introducing an unseen drawing type would require retraining the model. This limitation is considered acceptable as the well-known dataset in this field only consists of these drawings.

\item[(iii)] \textit{Population Bias}: The dataset might not comprehensively represent the full diversity of individuals with or without PD. If most participants in the dataset share similar characteristics such as age, handedness, the model might capture subgroup-specific patterns rather than the generalizable PD indicators. However, CV5-Img and LOIO are used to simulate these situations.

\end{itemize}

\section{Conclusion}\label{sec:conclusion}
We presented a new approach for Parkinson’s disease detection from hand-drawing tasks that combines drawing-type recognition with chunk-based feature extraction and classical machine learning classifiers. Using individual-wise evaluation, the presented approach achieved more stable accuracy while reducing the generalization gap observed in other works. Specifically, our method achieved \textbf{97.08\%} (Img-CV5), \textbf{94.91\%} (Ind-CV5), and \textbf{92.40\%} (LOIO), reducing the image-to-subject gap to \textbf{2.17} pp (and \textbf{4.68} pp to LOIO), compared with much larger drops reported by end-to-end baselines and the state-of-the-art. The Ablation study confirmed that \textbf{2$\times$2 chunking} and \textbf{deterministic augmentation} are both critical: chunking allows full representational capacity to localized strokes, enhancing sensitivity to fine motor irregularities, while augmentation increases and diversifies the training data, improving individual-wise robustness.


\bibliography{sn-article}

\end{document}